# Quantifying Overfitting: Introducing the Overfitting Index

*Sanad Aburass*
*Department of Computer Science*
*Maharishi International University*
*Fairfield, Iowa, USA*
*saburass@miu.edu.*

***Abstract*-In the rapidly evolving domain of machine learning, ensuring model generalizability remains a quintessential challenge. Overfitting, where a model exhibits superior performance on training data but falters on unseen data, is a recurrent concern. This paper introduces the Overfitting Index (OI), a novel metric devised to quantitatively assess a model's tendency to overfit. Through extensive experiments on the Breast Ultrasound Images Dataset (BUS) and the MNIST dataset using architectures such as MobileNet, U-Net, ResNet, Darknet, and ViT-32, we illustrate the utility and discernment of the OI. Our results underscore the variable overfitting behaviors across architectures and highlight the mitigative impact of data augmentation, especially on smaller and more specialized datasets. The ViT-32's performance on MNIST further emphasizes the robustness of certain models and the dataset's comprehensive nature. By providing an objective lens to gauge overfitting, the OI offers a promising avenue to advance model optimization and ensure real-world efficacy.**

***Keywords: Overfitting, Supervised Learning, Machine Learning***

## I. INTRODUCTION

In recent years, machine learning has transformed the way we process, analyze, and utilize data. Within the expansive realm of machine learning, supervised learning remains one of its most potent subfields [1]. Supervised learning involves training a model on a set of input-output pairs, aiming to find a general rule that maps inputs to outputs [2]. As the model learns from the training data, it aspires to make accurate predictions on unseen or new data [3]. Among the numerous methodologies within supervised learning, deep learning - characterized by neural networks with multiple layers - has proven especially influential [4]. Its capabilities in handling vast data sets and intricate patterns have led to breakthroughs in various applications, from computer vision to natural language processing. However, the profound capacities of deep learning come with their own set of challenges [5]. One such issue, which has been central to the discourse within the machine learning community, is that of 'overfitting'. Overfitting occurs when a model learns the training data too closely, capturing even its noise and outliers, which results in poor generalization to new, unseen data [6]. In essence, an overfitted model has become overly complex, tailoring itself excessively to the training set, and thus, performs poorly on validation or test sets [7]. Overfitting is not just a theoretical concern but has practical implications [8]. Models that overfit are often unreliable when deployed in real-world scenarios since they might not perform as expected on new data [9]. As such, researchers and practitioners are in constant pursuit of techniques to detect, prevent, and mitigate overfitting [10]. Existing techniques and metrics primarily focus on monitoring the performance gap between the training and validation sets [11]. However, these might not provide a holistic view of overfitting as it unfolds over the training process, particularly over epochs in the context of deep learning [12]. There's a clear need for a more intuitive, comprehensive metric that encapsulates the dynamics of overfitting over time, considering both loss and accuracy. Enter the Overfitting Index (OI). This paper introduces the OI, a novel metric designed to provide a clearer quantification of overfitting, considering its progression across epochs. By integrating the differences in loss and accuracy between training and validation sets, and weighting them by the epoch number, the OI offers a nuanced view of overfitting as it emerges and intensifies during training.

In light of the pivotal role of machine learning models in modern applications, it is imperative to ensure their reliability and robustness. By adopting the OI, practitioners and researchers can gain more insightful diagnostics into their models, fostering the development of models that generalize better and yield trustworthy results. As we delve deeper into this paper, we will elucidate the intricacies of the OI, demonstrating its potential as a staple metric in machine learning model evaluation.

## II. PROPOSED APPROACH

### *A. Understanding Overfitting*

At the heart of our proposed approach lies the phenomenon of overfitting. Overfitting occurs when a machine learning model performs exceptionally well on the training data but fails to generalize effectively to new, unseen data. In essence, it indicates that the model has learned the peculiarities, noise, and outliers of the training data, rather than capturing the underlying patterns that are broadly applicable. The primary indication of overfitting can be observed in the disparity between a model's performance on its training and validation datasets. Typically, as a model is trained over epochs, its accuracy on the training data will increase, and its associated loss will decrease. However, if overfitting sets in, the model's performance on the validation data will begin to degrade after a certain point, even if its training performance continues to improve [9]–[12].

### B. The Overfitting Index Equation

Given the criticality of overfitting and its implications, we introduce the Overfitting Index (OI) to measure the degree of overfitting across the model's training period. The equation is as follows:

$$\boldsymbol{OI} = \sum_{e=1}^{N} \mathbf{max}(\boldsymbol{Loss\ Diffreance}, \boldsymbol{Accuracy\ Diffrence}) * \boldsymbol{e} \quad (1)$$

Where:
- e represents the epoch number.
- N is the total number of epochs.
- Loss Difference is calculated as (Validation Loss - Training Loss) but only considered when Validation Loss > Training Loss.
- Accuracy Difference is calculated as (Training Accuracy - Validation Accuracy) but only when Training Accuracy > Validation Accuracy.

The essence of the OI formula lies in capturing the most prominent difference between the validation and training datasets, whether it's in terms of loss or accuracy. By multiplying this difference with the epoch number, we provide a weighted measure, emphasizing more pronounced overfitting that occurs in later epochs. This approach ensures that our metric doesn't just capture a snapshot of overfitting at a particular epoch but provides an aggregated measure over the course of the training. Therefore, OI gives a holistic representation of how the model's propensity to overfit evolves as training progresses.

### C. Mitigating Overfitting

Recognizing overfitting is only the first step. The subsequent and equally significant challenge lies in its mitigation. Several strategies can be employed:

1. Regularization Techniques: Methods such as L1 and L2 regularization can prevent the coefficients of the model from becoming too large, thus ensuring the model doesn't become overly complex.

2. Early Stopping: By monitoring the model's performance on a validation dataset, training can be halted as soon as the performance starts deteriorating, indicating the onset of overfitting.

3. Dropout: Especially useful in deep learning, dropout involves randomly setting a fraction of input units to 0 at each update during training, which can prevent over-reliance on any particular neuron[6].

4. Data Augmentation: Increasing the diversity of training data by applying various transformations can make the model more robust and less likely to overfit [13]–[15].

5. Pruning: In neural networks, pruning involves removing certain neurons or connections, thus simplifying the model.

With the introduction of the OI, practitioners now have an additional tool to monitor and diagnose overfitting. Coupling this metric with the aforementioned mitigation techniques offers a comprehensive strategy to combat overfitting, ensuring models are both accurate and generalizable.

## III. EXPERIMENTAL RESULTS

### A. Dataset and Setup

To empirically validate the efficacy of our proposed OI, we employed two distinct datasets:

1. Breast Ultrasound Images Dataset: Sourced from Kaggle, this dataset offers a collection of breast ultrasound images. Medical imaging datasets, including the one we selected, often suffer from overfitting. This is primarily because medical datasets tend to be smaller in size compared to other domains, due to privacy concerns, challenges in data collection, and the specialized nature of the data. Consequently, the inherent complexity and variability of such datasets make them a challenging task for most machine learning models.

2. MNIST Dataset: A staple in the machine learning community, the MNIST dataset comprises hand-written digits. Given its vast size and relative simplicity, models typically do not overfit on this dataset, providing an interesting contrast to the medical imaging dataset in our study.

All experiments were performed on Google Colab Pro, capitalizing on the computational efficiency of the T4 GPU. This ensured both consistent model performance and efficient training times.

### B. Models, Training Paradigm, and Dataset Specifics

For the Breast Ultrasound Images Dataset, we tested four contemporary architectures:

1. MobileNet: A streamlined model optimized for mobile and embedded vision applications.
2. U-Net: Specifically designed for biomedical image segmentation, its architecture both contracts and then expands.
3. ResNet: Renowned for its unique residual blocks, which facilitate the training of deep neural networks by addressing the vanishing gradient problem.
4. Darknet: Primarily associated with its use in the YOLO (You Only Look Once) real-time object detection system.

Each of these models underwent two separate training regimens:

1. Without Data Augmentation: Initially, models were trained directly on the original dataset without any form of augmentation.
2. With Data Augmentation: The dataset was subsequently augmented to introduce variability. The models were then trained again on this expanded dataset to assess any changes in their performance and overfitting behavior.

For the MNIST Dataset, the ViT-32 (Vision Transformer with a patch size of 32x32) model was trained. Given MNIST's comprehensive nature, the focus was on discerning the OI for a model trained on a dataset that is historically known to resist overfitting, especially without the need for data augmentation.

### *C. Overfitting Index Evaluation*

Upon the conclusion of training, the OI was computed for each model under their respective training conditions. The goal was to elucidate the tendency of each model to overfit, especially when exposed to diverse datasets and varying training methodologies. Table 1, alongside Figures 1 through 18, presents our experimental outcomes. We've graphically represented the trends observed during the final 10 epochs for each model, focusing on the contrasts between training and validation metrics — both in terms of loss and accuracy. A consistent observation across most experiments is the superior performance of training metrics compared to their validation counterparts. Specifically, the training loss typically demonstrates a steady decrease while the training accuracy showcases a consistent rise, often reaching a point where they seem to plateau, indicating convergence. In contrast, the validation metrics tend to exhibit more volatility. The validation loss and accuracy often oscillate, rather than following a steady trend, indicating potential overfitting.
However, a notable exception is observed with the ViT-32 model trained on the MNIST dataset. Here, the OI is notably low, and the plots illustrate a stable trend in both validation loss and accuracy, hinting at a well-generalized model performance.

TABLE I: OI Scores for the Models

| **Model** | **OI (Without Augmentation)** | **OI (With Augmentation)** |
|---|---|---|
| MobileNet on BUS | 6531.3628 | 3819.9278 |
| U-Net on BUS | 337.866 | 195.7435 |
| ResNet on BUS | 496.149 | 388.6641 |
| Darknet on BUS | 2774.4355 | 650.3336 |
| ViT32 on MNIST | **2.0354** | N/A |

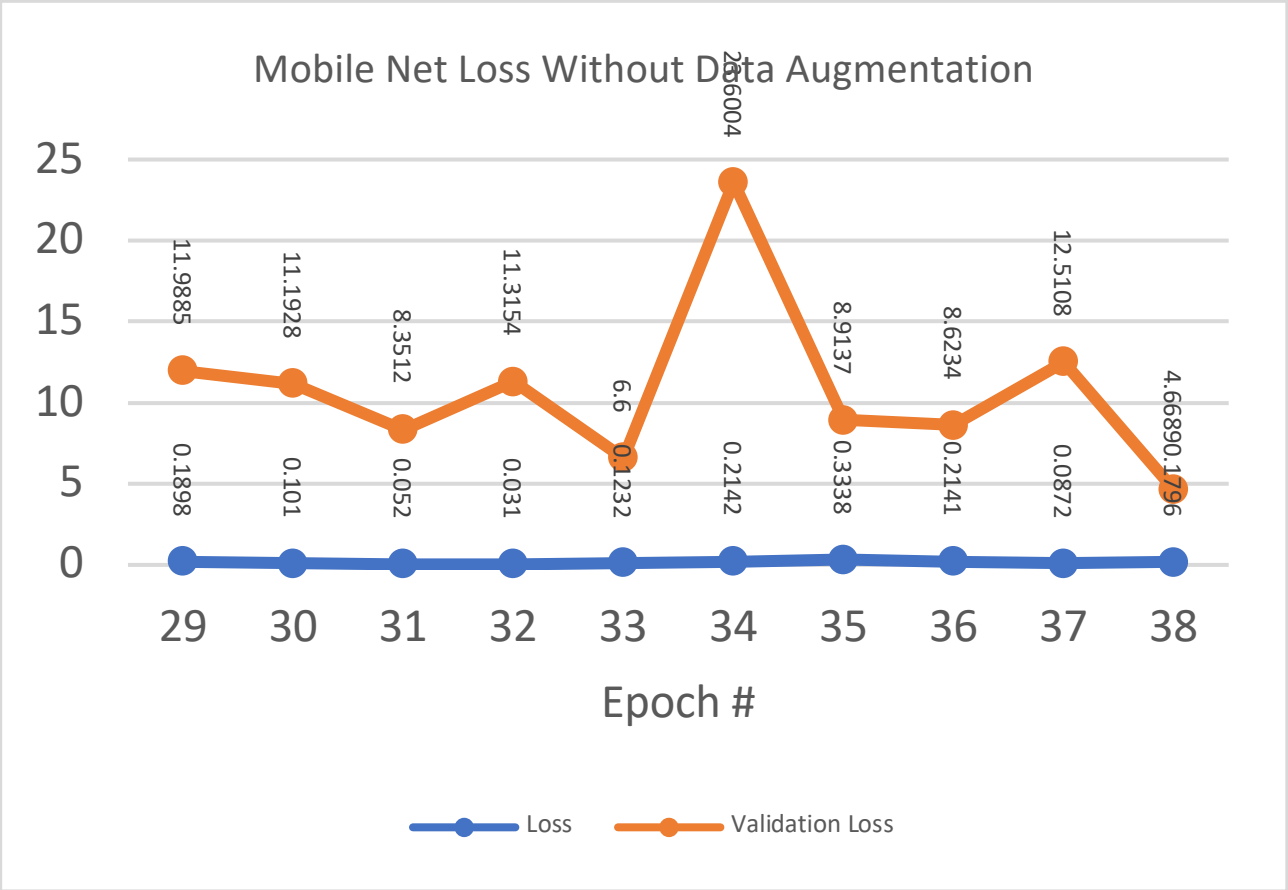

Fig 1. Mobile Net Loss Without Data Augmentation

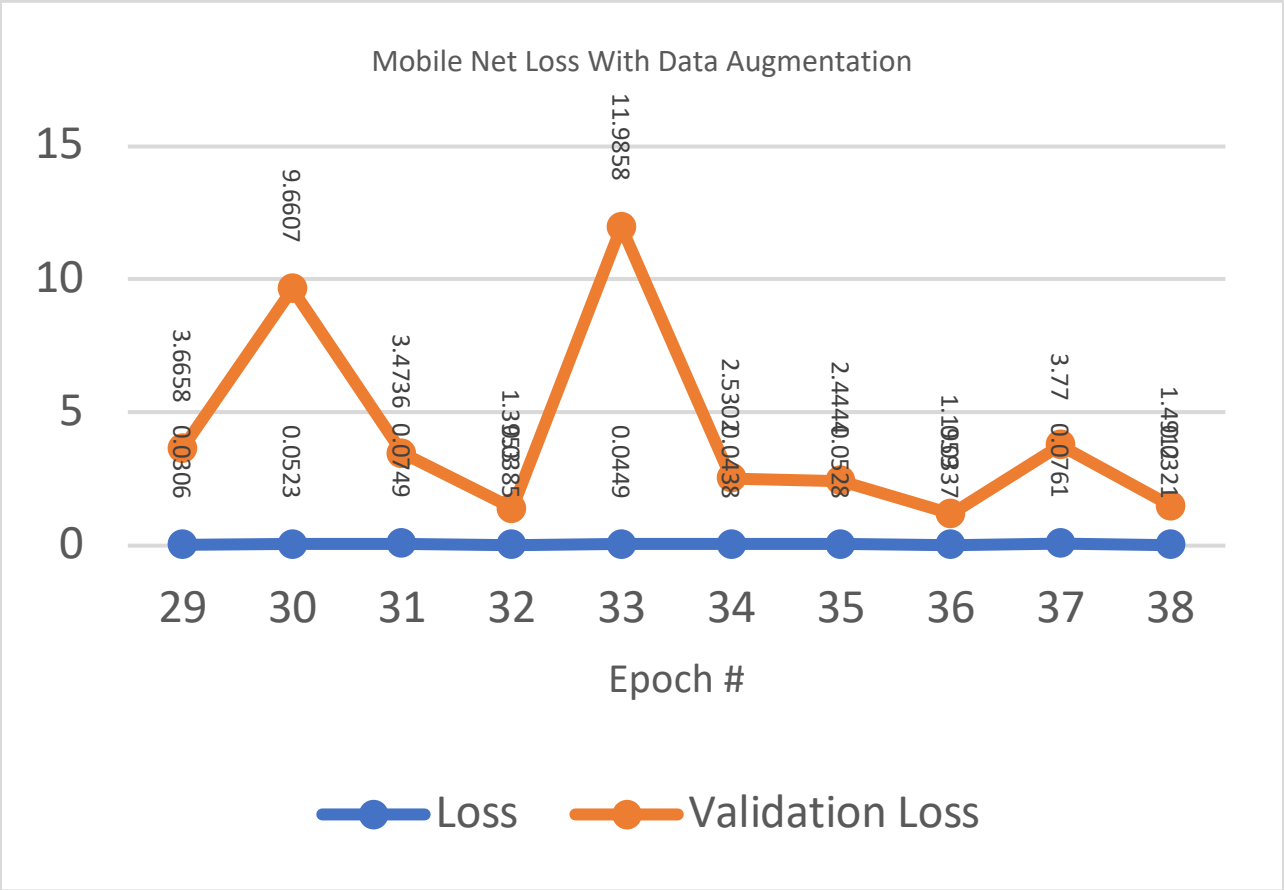

Fig 3. Mobile Net Loss With Data Augmentation

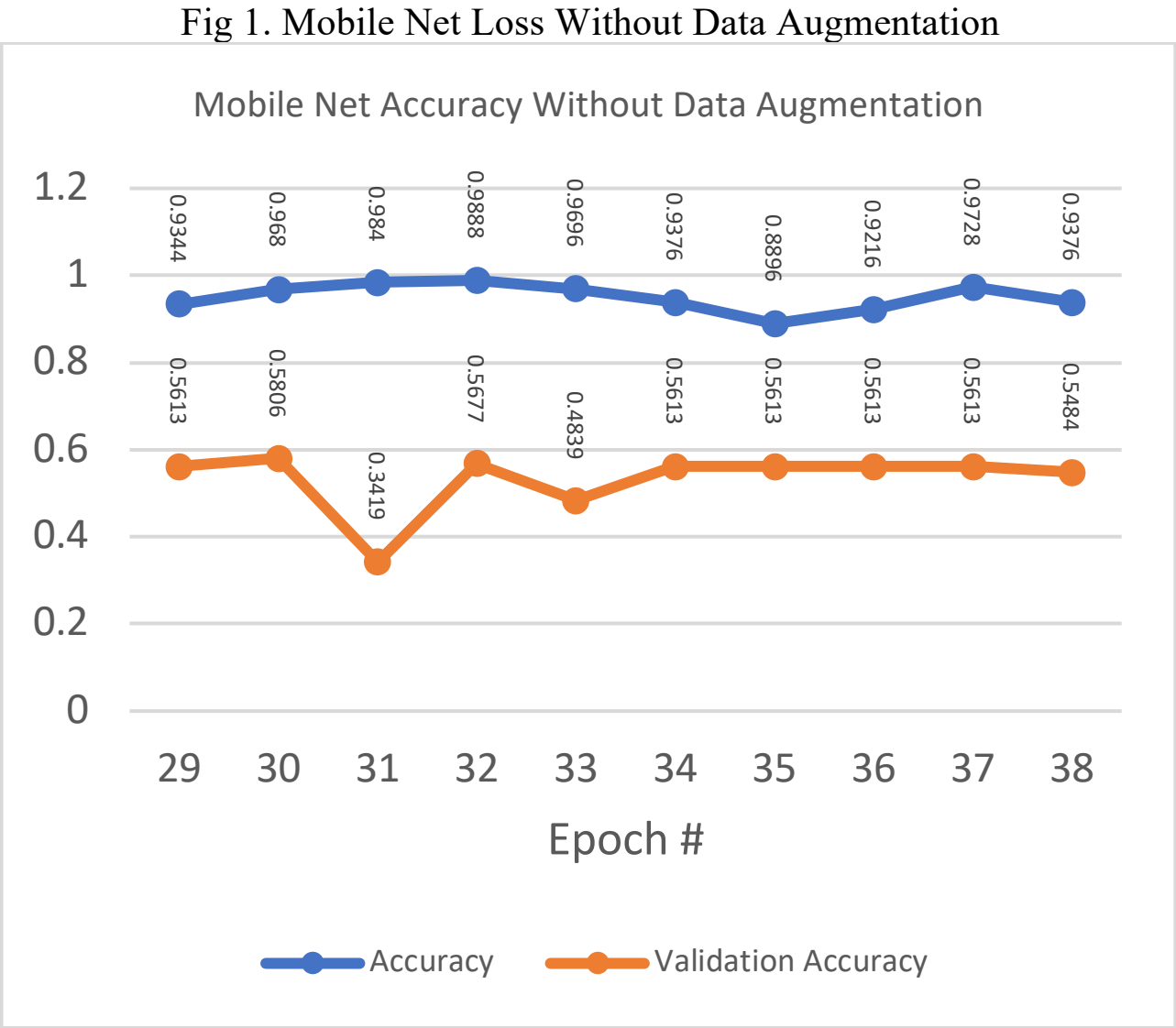

Fig 2. Mobile Net Accuracy Without Data Augmentation

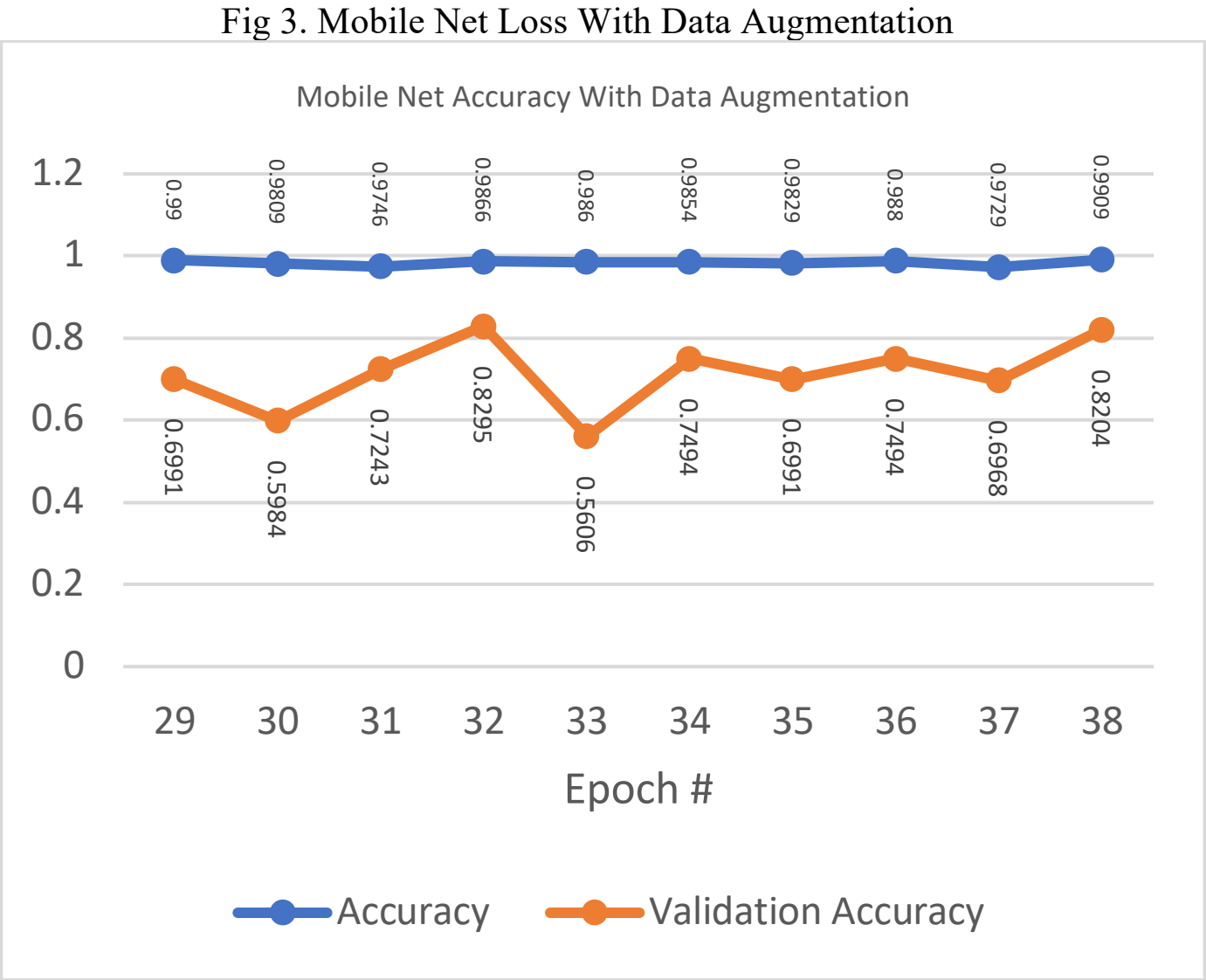

Fig 4. Mobile Net Accuracy With Data Augmentation

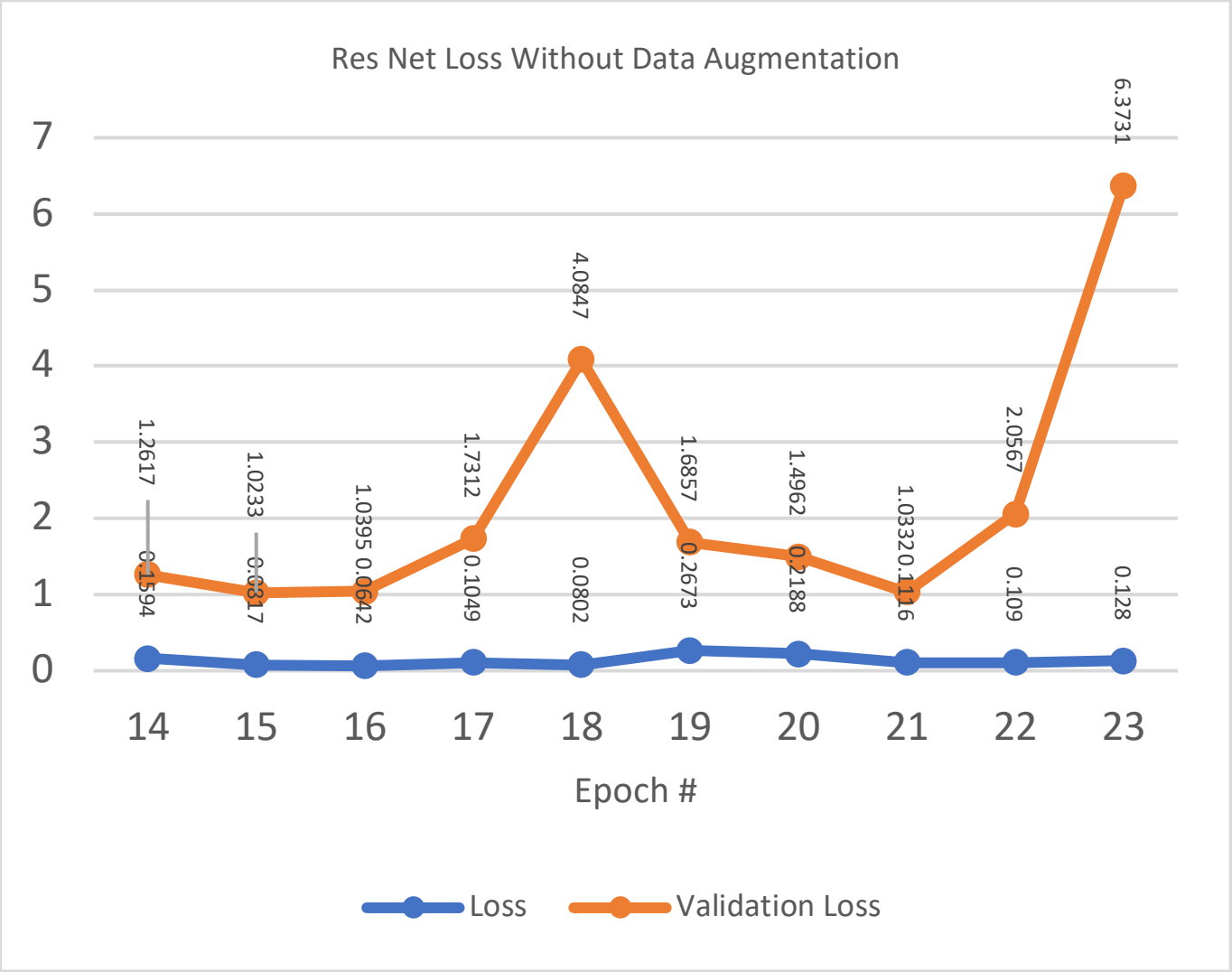

Fig 5. Res Net Loss Without Data Augmentation

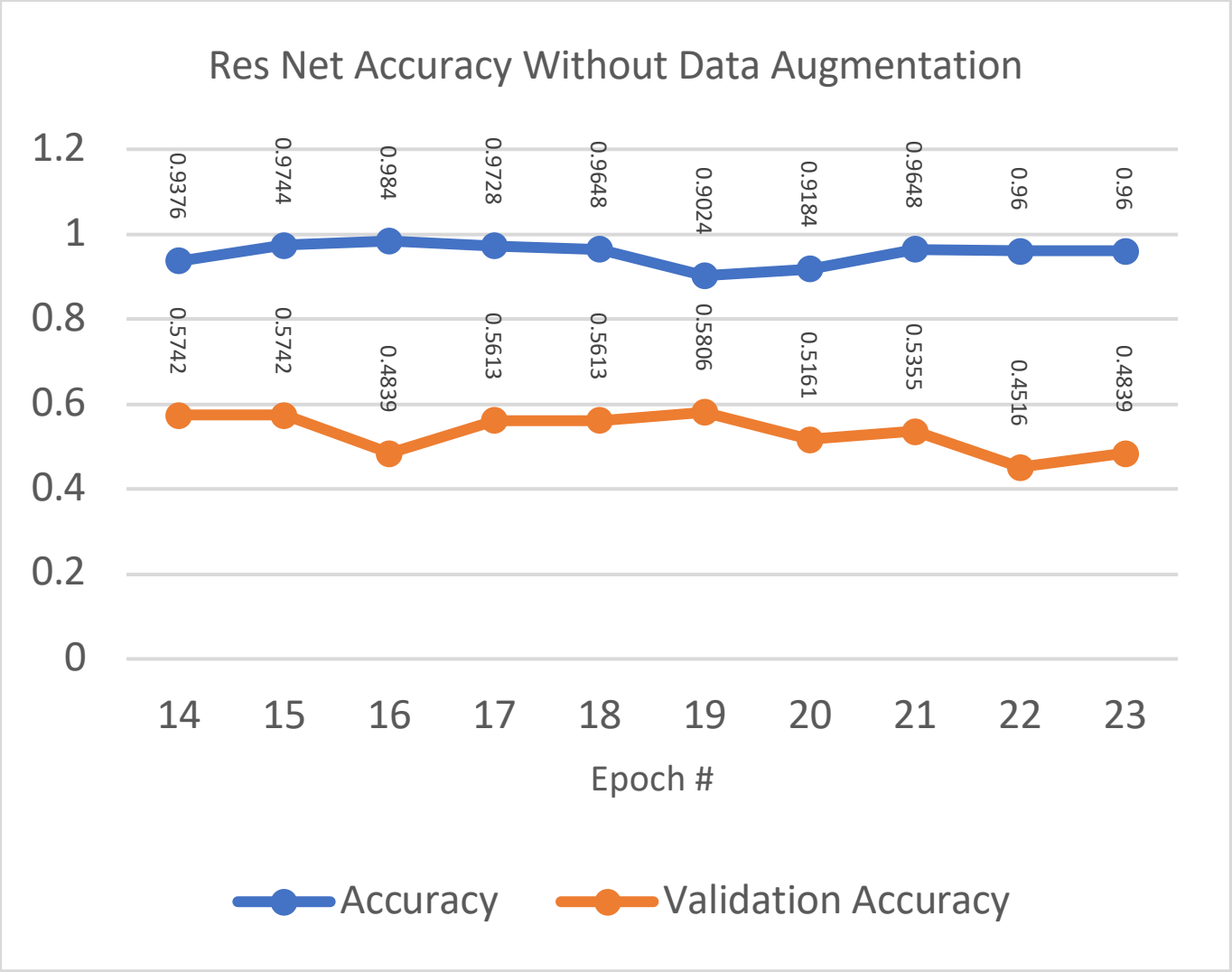

Fig 6. Res Net Accuracy Without Data Augmentation

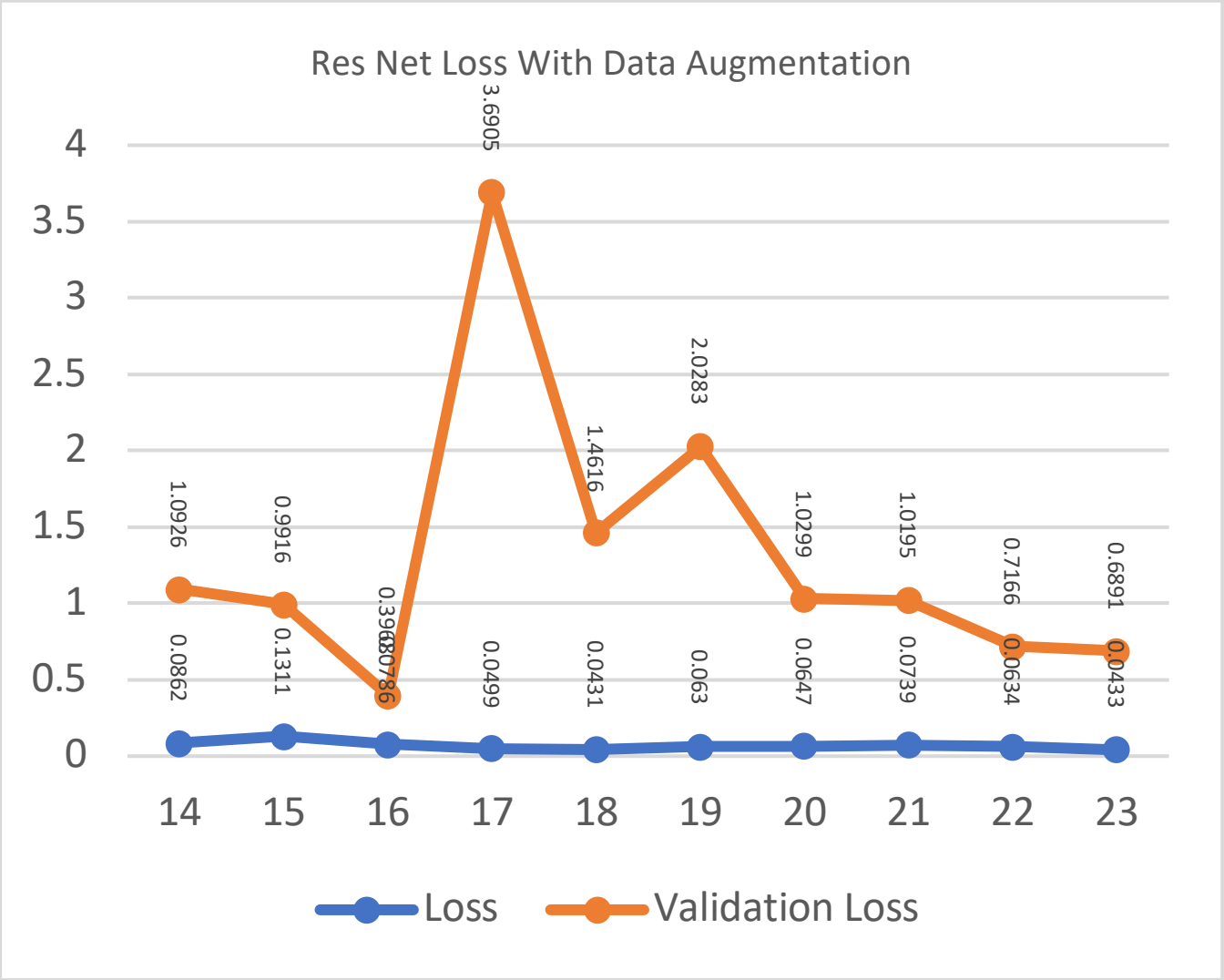

Fig 7. Res Net Loss With Data Augmentation

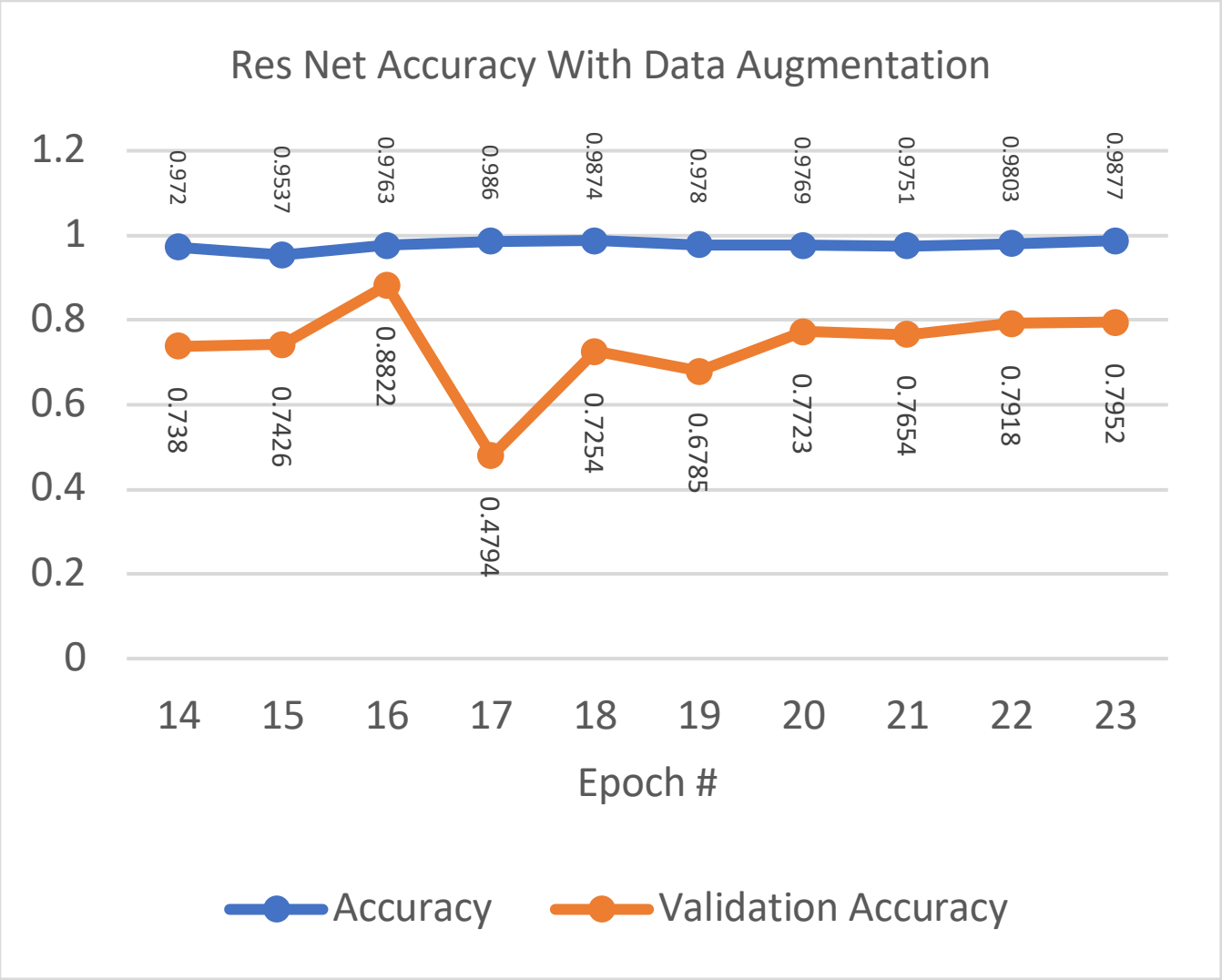

Fig 8. Res Net Accuracy With Data Augmentation

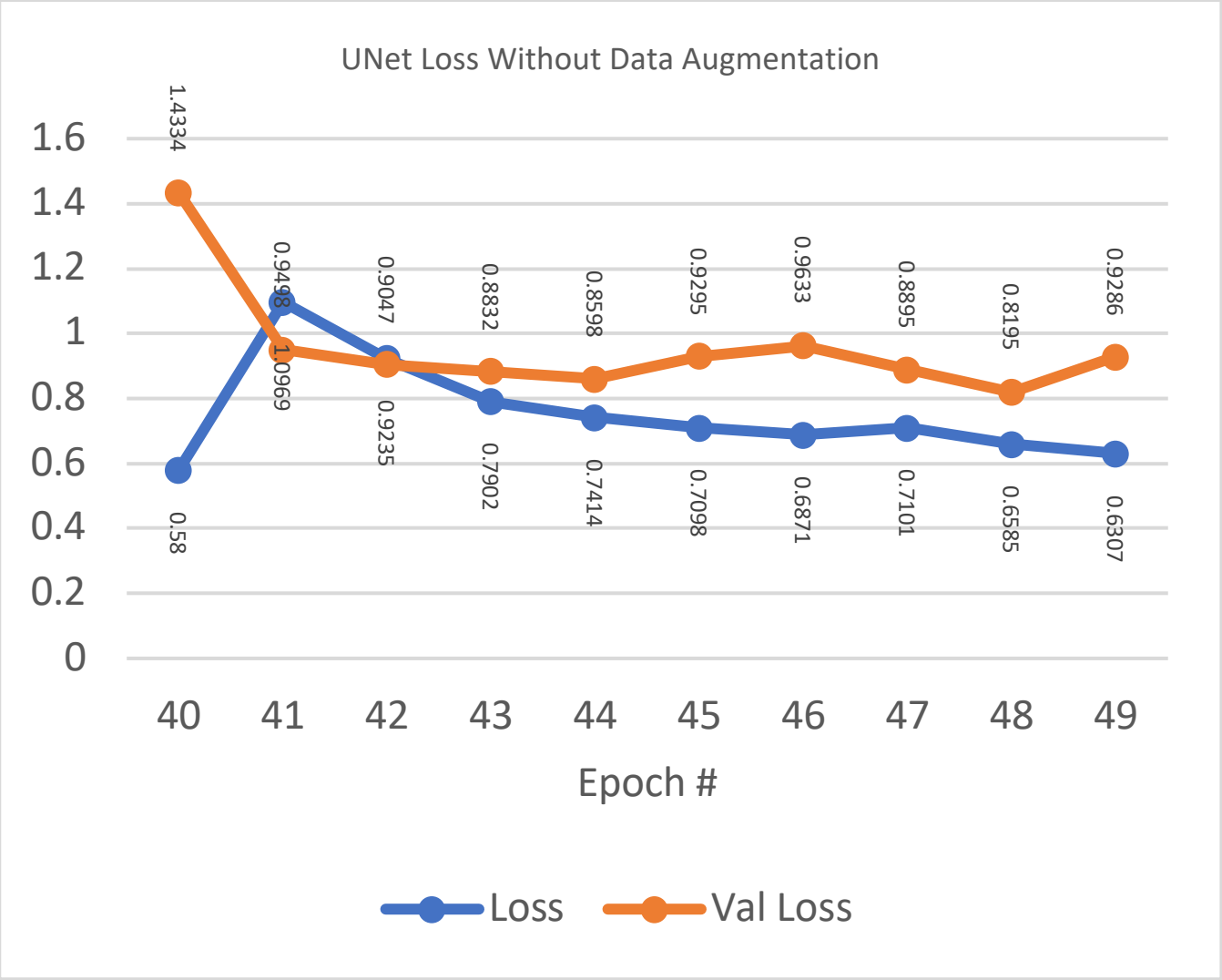

Fig 9. UNet Loss Without Data Augmentation

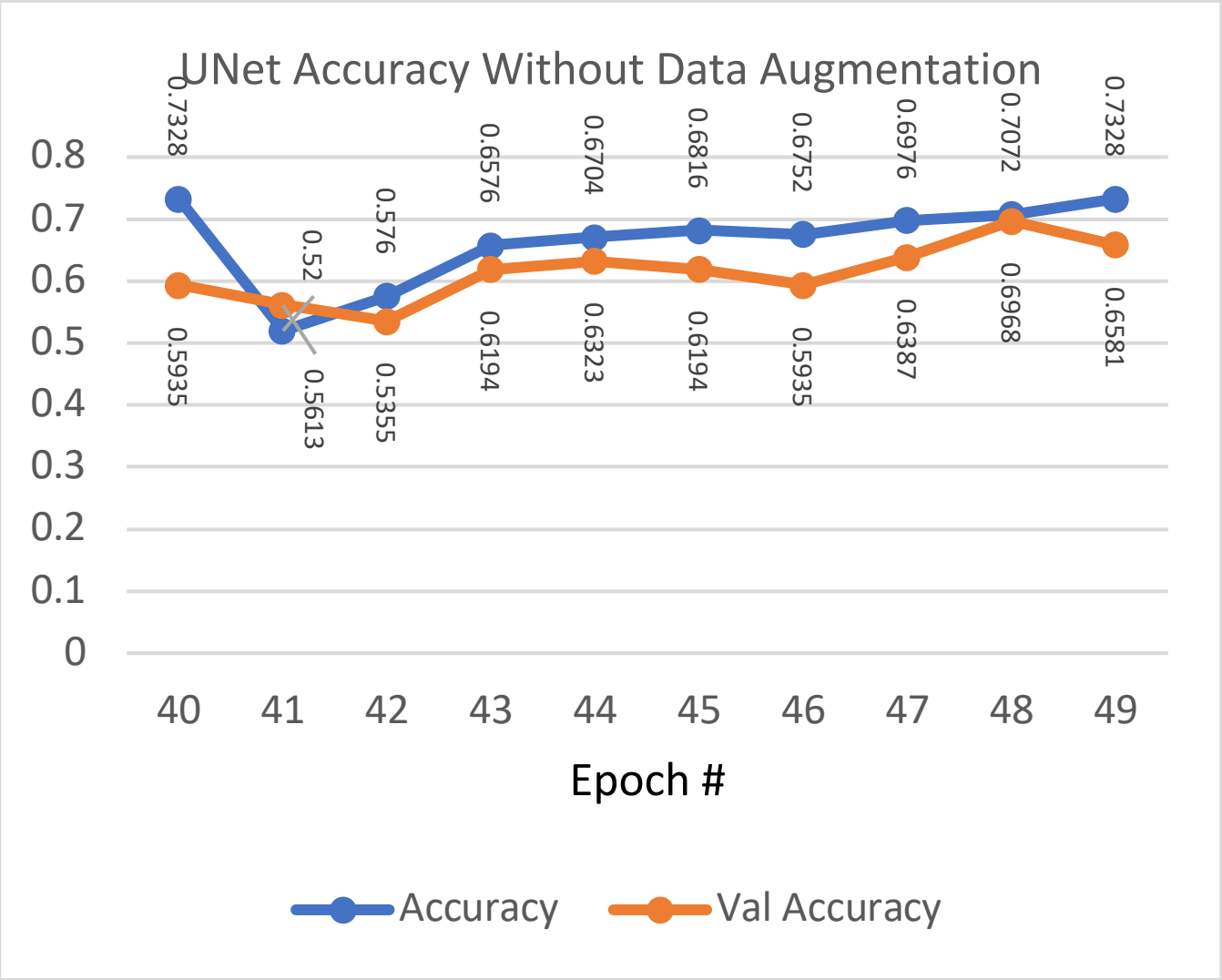

Fig 10. UNet Accuracy Without Data Augmentation

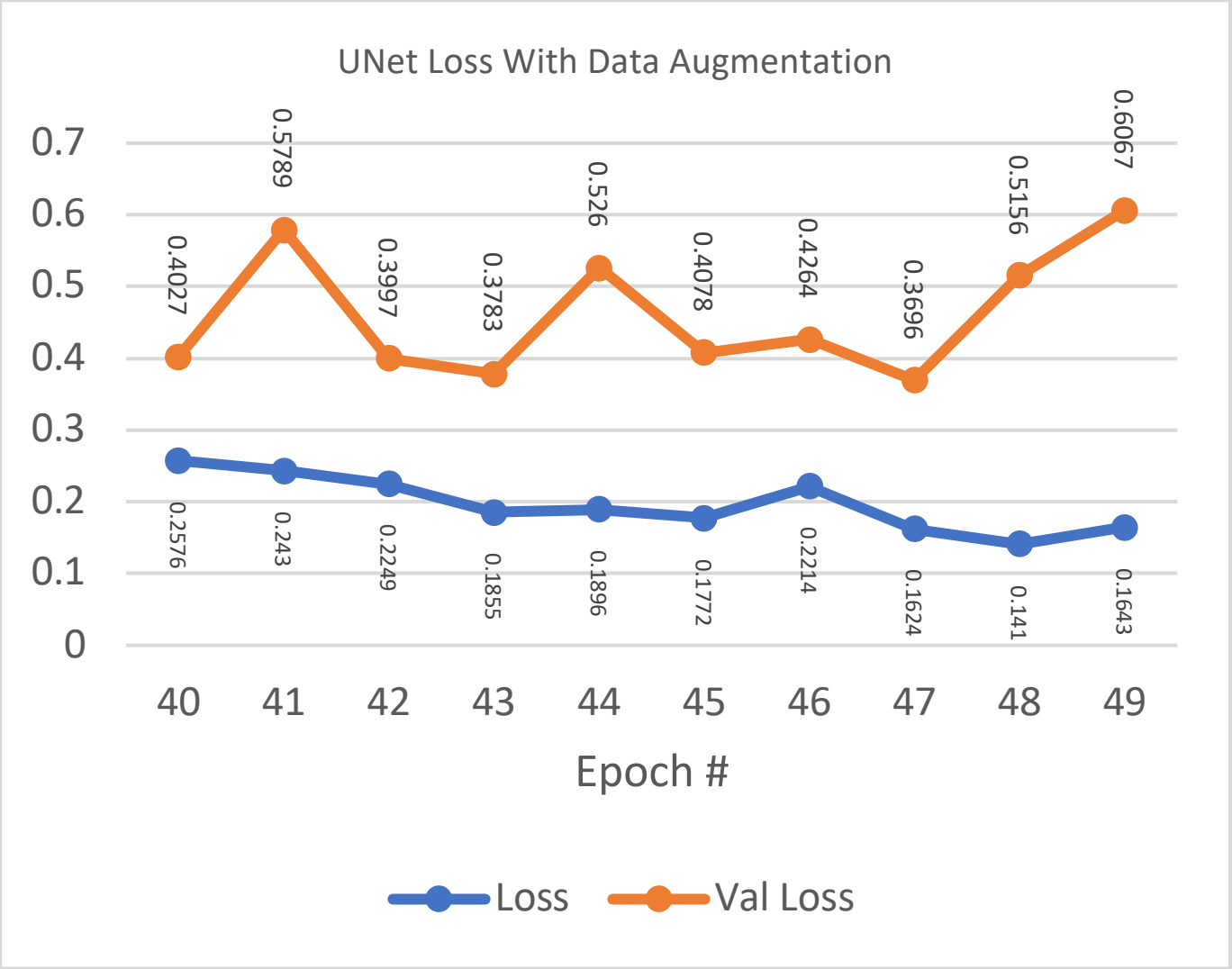

Fig 11. UNet Loss With Data Augmentation

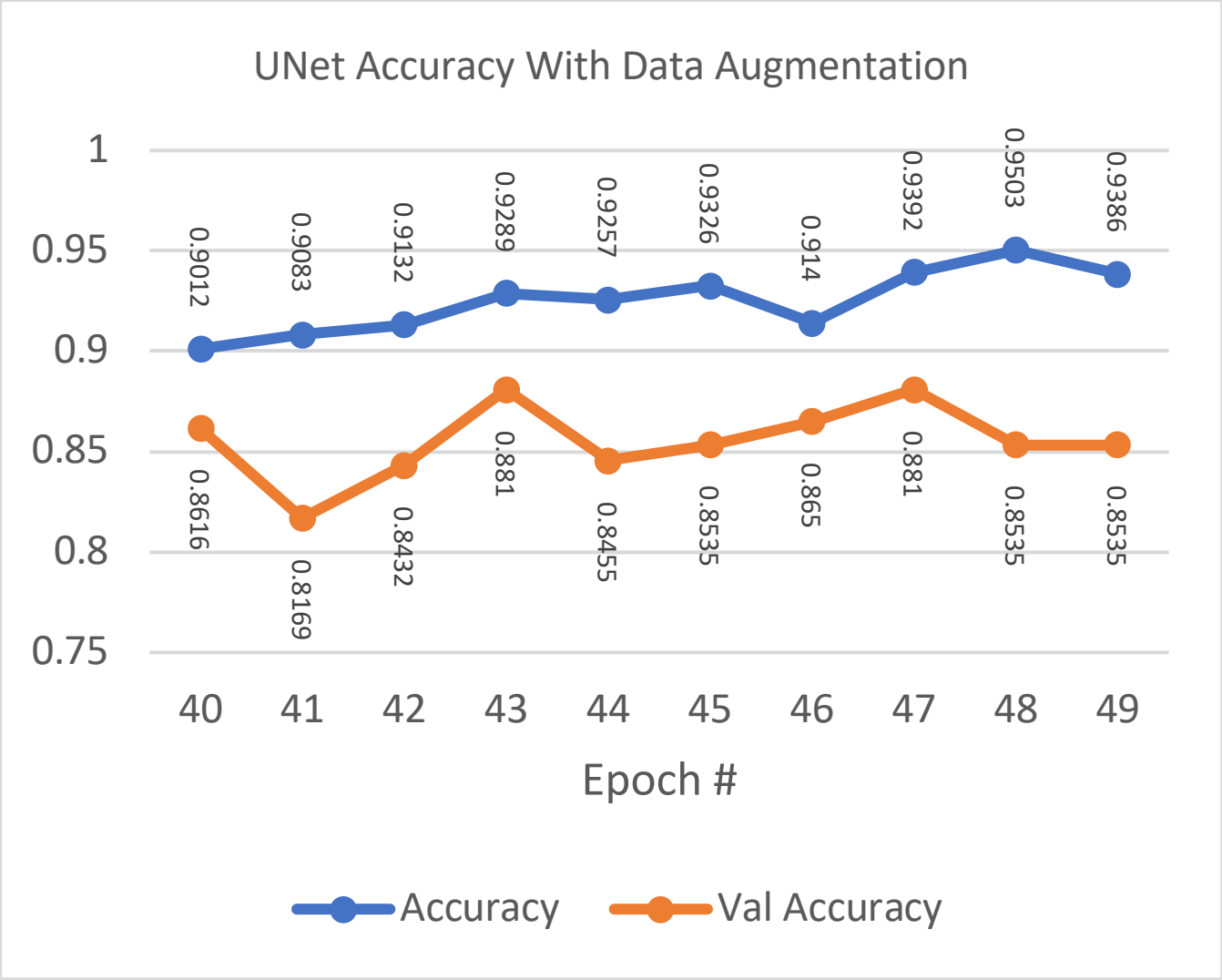

Fig 12. UNet Accuracy With Data Augmentation

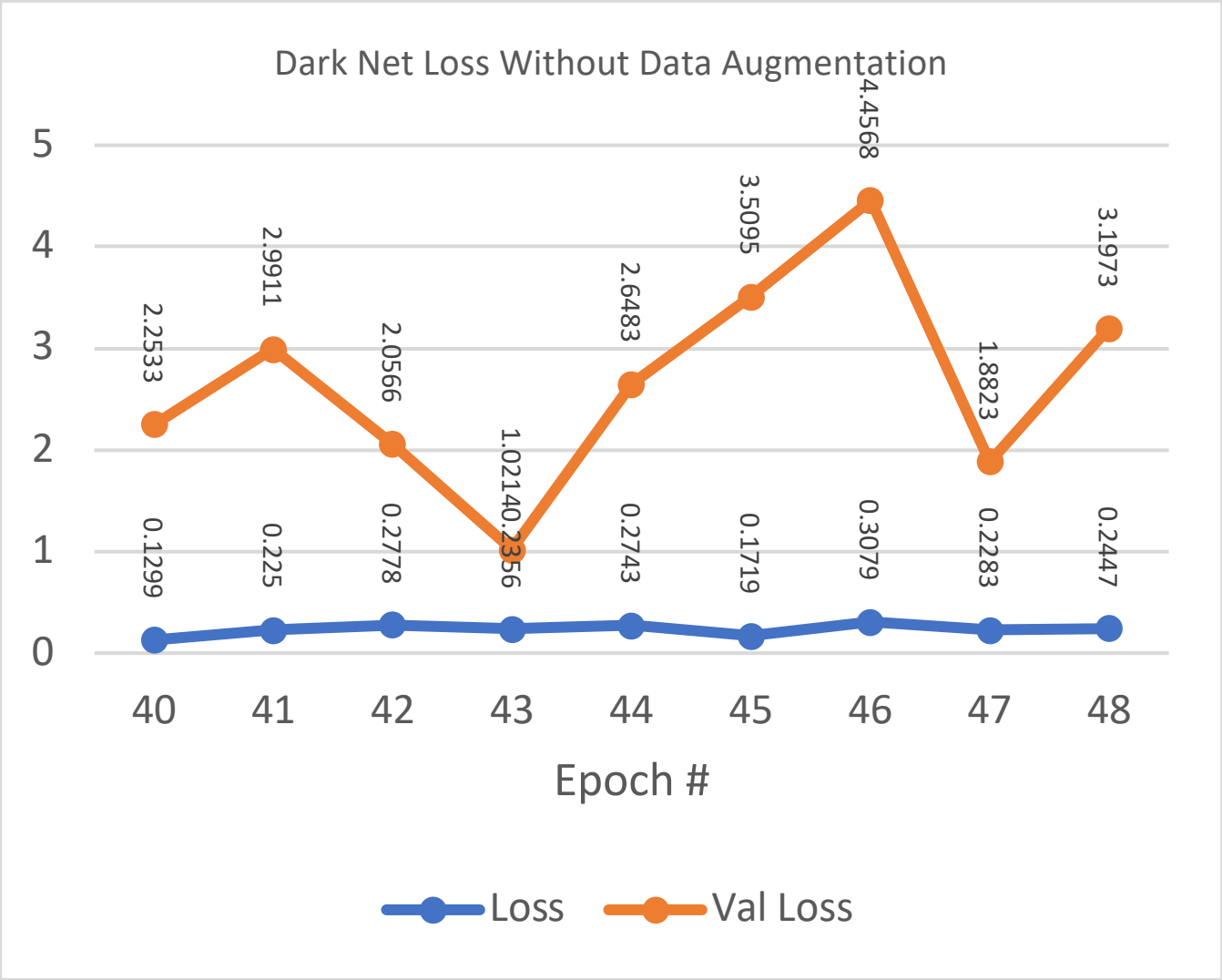

Fig 13. Dark Net Loss Without Data Augmentation

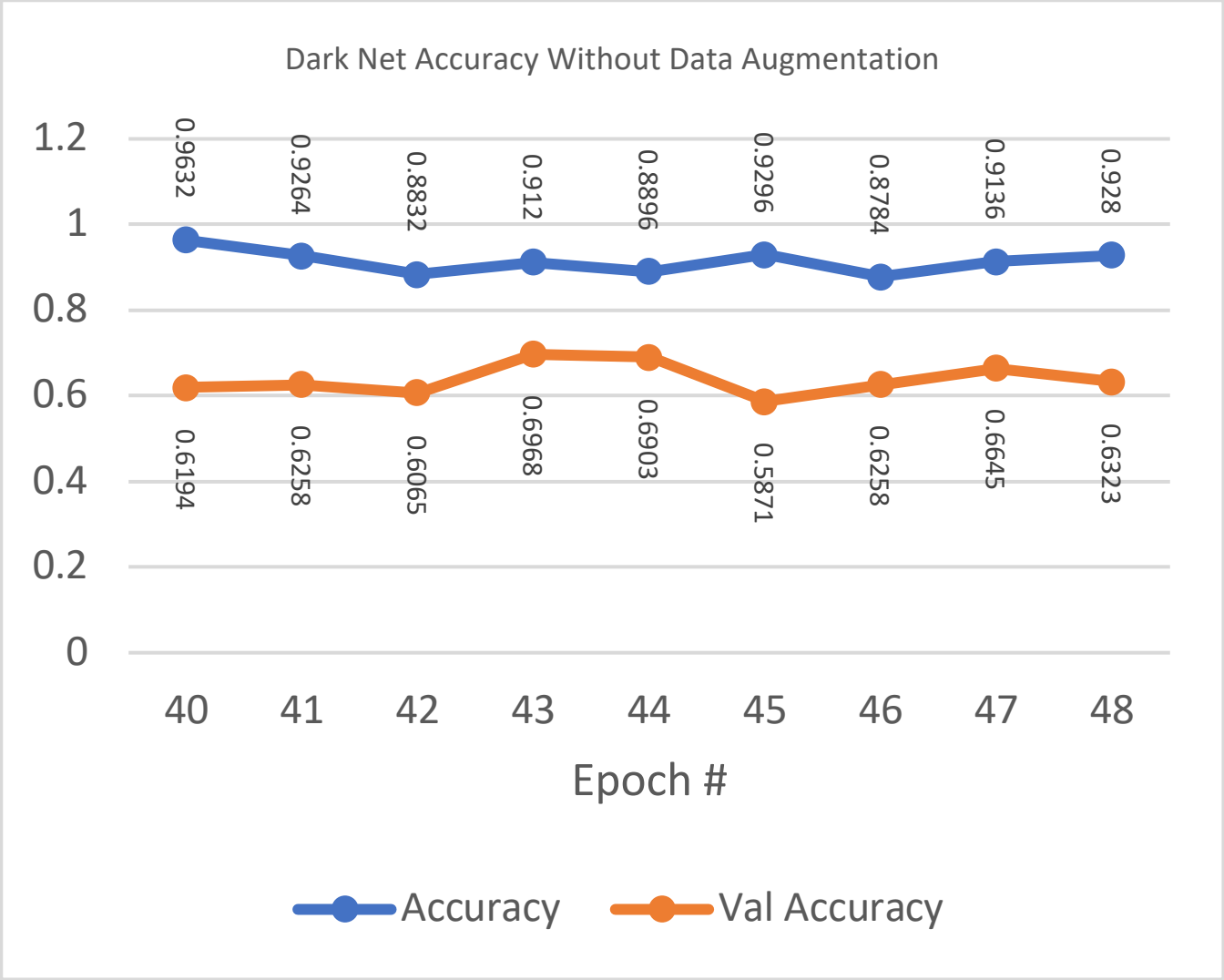

Fig 14. Dark Net Accuracy Without Data Augmentation

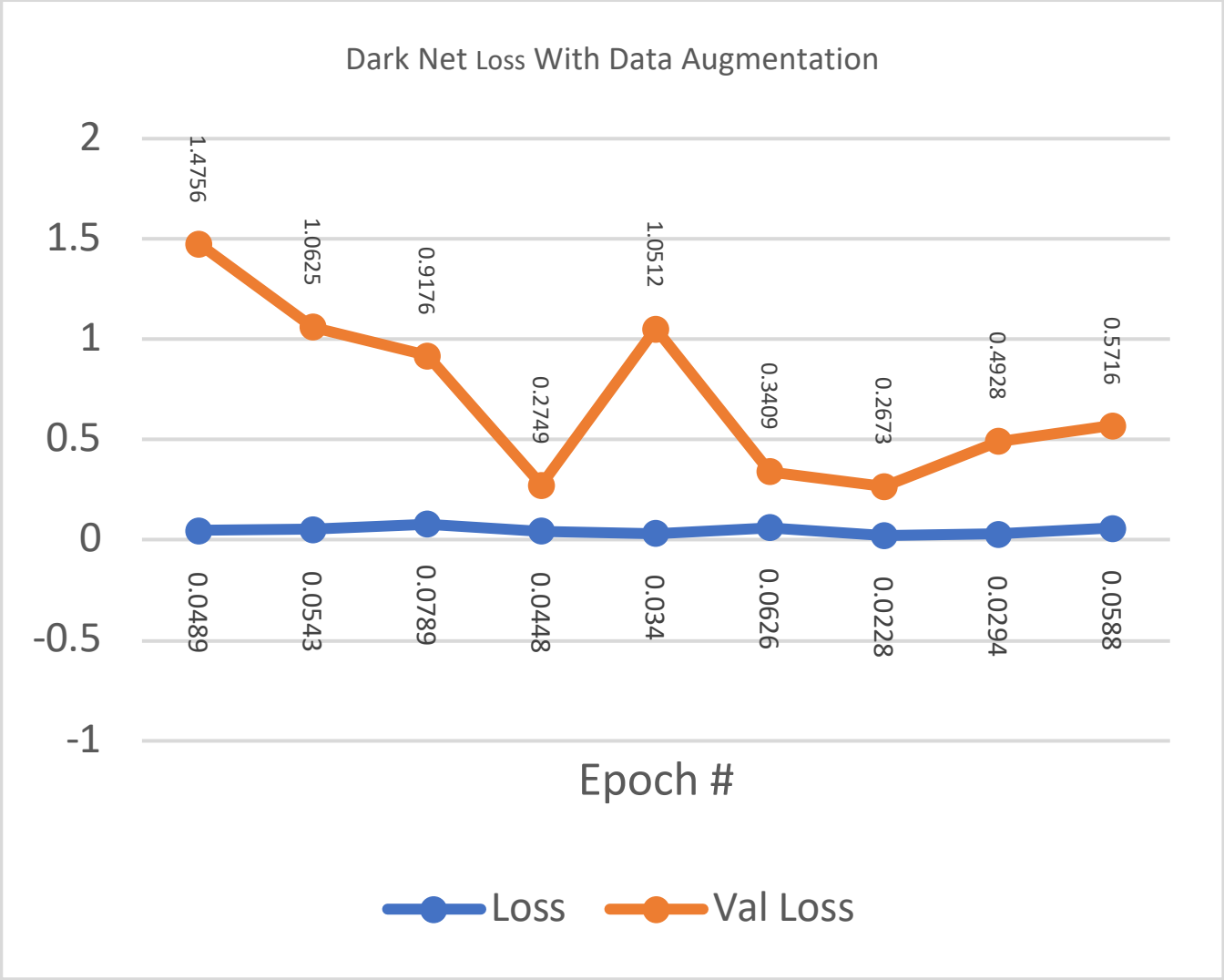

Fig 15. Dark Net Loss With Data Augmentation

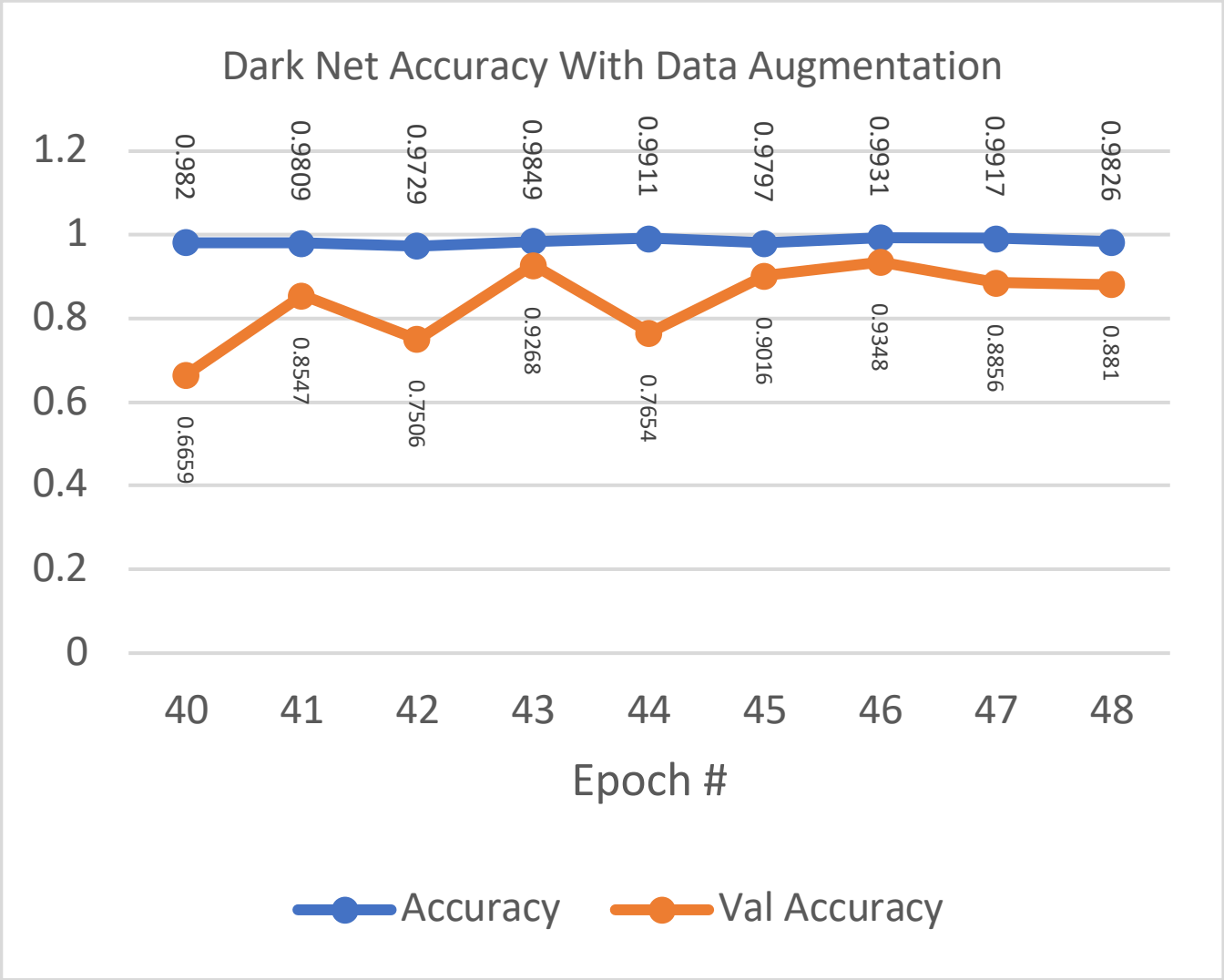

Fig 16. Dark Net Accuracy With Data Augmentation

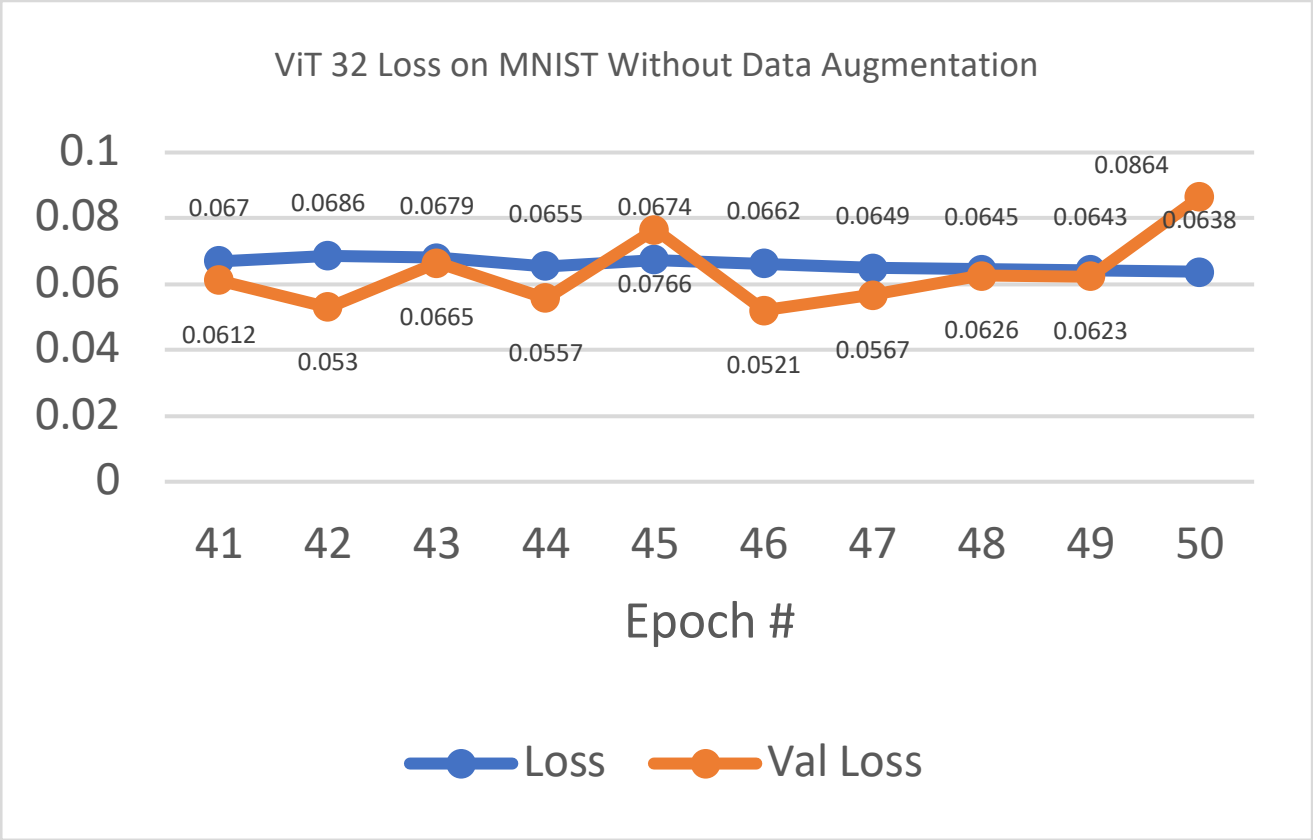

Fig 17. ViT 32 Loss on MNIST Without Data Augmentation

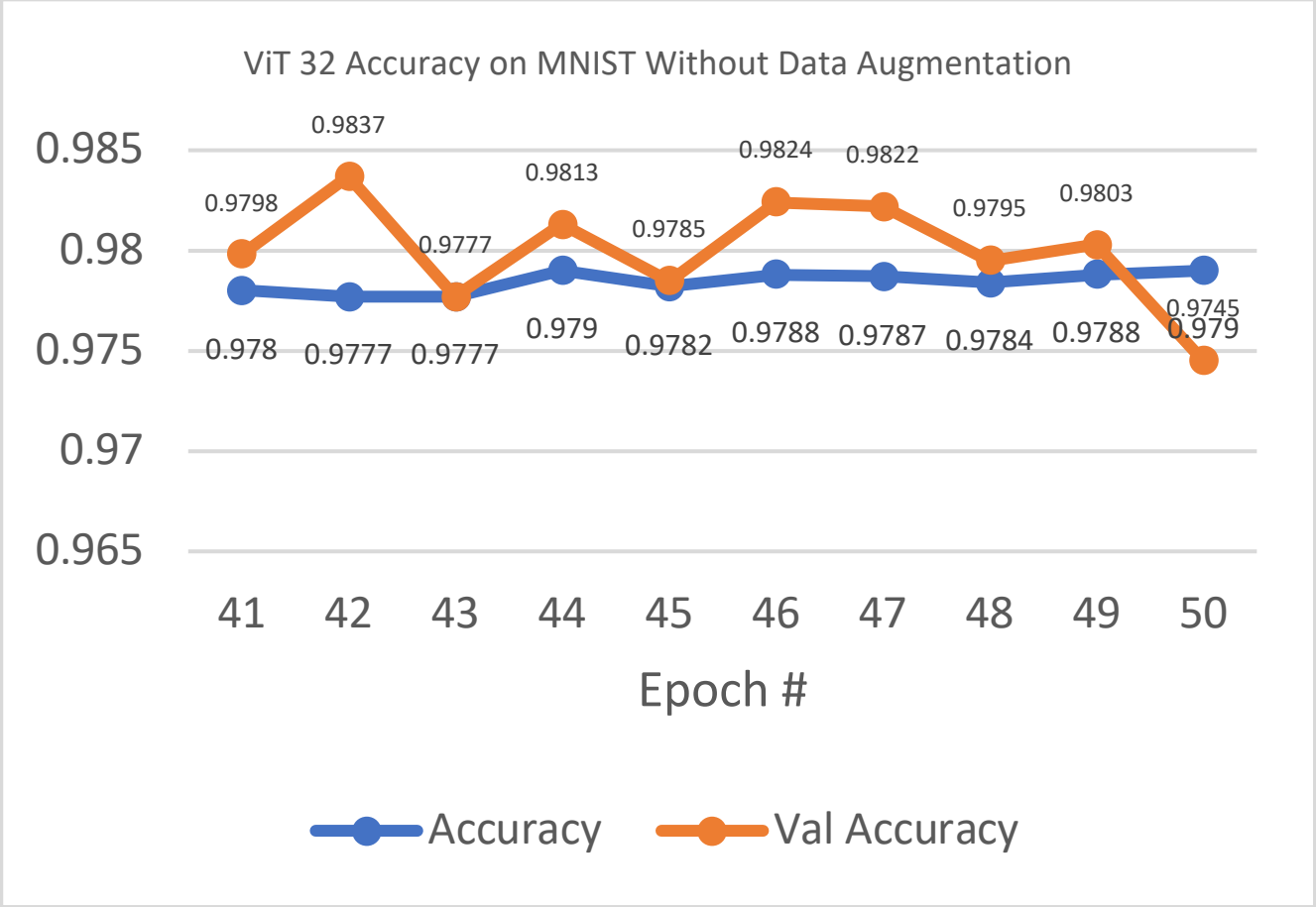

Fig 18. ViT 32 Accuracy on MNIST Without Data Augmentation

## IV. DISCUSSION

In our study, the OI has emerged as a pivotal metric, providing quantifiable insights into a model's propensity to overfit across various datasets and architectures. Notably, MobileNet and Darknet on the Breast Ultrasound Images Dataset (BUS) displayed significant susceptibility to overfitting, but this was notably reduced with data augmentation. In contrast, architectures like U-Net and ResNet demonstrated inherent regularization properties, exhibiting relatively lower OI values. The near-minimal OI of ViT-32 on MNIST accentuated the dataset's comprehensive nature and the model's robustness. Such discerning results underline OI's efficacy, emphasizing its potential as an invaluable tool. This unified metric not only offers objective evaluations of model behavior across datasets but also serves as a compass guiding model optimization, selection, and the effective deployment of data augmentation strategies.

## V. CONCLUSION

As the realms of machine learning and deep learning continue to grow, the challenge of model overfitting stands as a persistent barrier to achieving universally effective and reliable model generalization. This paper's introduction of the OI has provided a robust, quantifiable metric to assess and understand the intricacies of overfitting across varied datasets and architectures. Our extensive empirical evaluations using both the Breast Ultrasound Images Dataset (BUS) and the MNIST dataset have highlighted the diverse behaviors of contemporary models. These experiments underscored the importance of strategic interventions, such as data augmentation, especially when navigating the nuanced challenges posed by specialized datasets like medical images. Moreover, the resilient performance of ViT-32 on MNIST serves as an optimistic reminder of the strides the field has made in building robust models. In essence, the OI stands poised to become an invaluable tool for researchers and practitioners alike, driving the next wave of innovations that not only perform well in controlled environments but also thrive in real-world scenarios.

## REFERENCES

[1] S. Aburass, A. Huneiti, and M. B. Al-Zoubi, “Classification of Transformed and Geometrically Distorted Images using Convolutional Neural Network,” *Journal of Computer Science*, vol. 18, no. 8, pp. 757–769, 2022, doi: 10.3844/jcssp.2022.757.769.

[2] N. Burkart and M. F. Huber, “A Survey on the Explainability of Supervised Machine Learning,” *Journal of Artificial Intelligence Research*, vol. 70, pp. 245–317, Jan. 2021, doi: 10.1613/jair.1.12228.

[3] Z. Li, F. Liu, W. Yang, S. Peng, and J. Zhou, “A Survey of Convolutional Neural Networks: Analysis, Applications, and Prospects,” *IEEE Trans Neural Netw Learn Syst*, vol. 33, no. 12, pp. 6999–7019, Dec. 2022, doi: 10.1109/TNNLS.2021.3084827.

[4] A. Khan, A. Sohail, U. Zahoora, and A. S. Qureshi, “A survey of the recent architectures of deep convolutional neural networks,” *Artif Intell Rev*, vol. 53, no. 8, pp. 5455–5516, Dec. 2020, doi: 10.1007/s10462-020-09825-6.

[5] S. Khan, M. Naseer, M. Hayat, S. W. Zamir, F. S. Khan, and M. Shah, “Transformers in Vision: A Survey,” *ACM Comput Surv*, vol. 54, no. 10s, pp. 1–41, Jan. 2022, doi: 10.1145/3505244.

[6] N. Srivastava, G. Hinton, A. Krizhevsky, I. Sutskever, and R. Salakhutdinov, “Dropout: A Simple Way to Prevent Neural Networks from Overfitting,” *J. Mach. Learn. Res.*, vol. 15, no. 1, pp. 1929–1958, Jan. 2014.

[7] S. Aburass and O. Dorgham, “Performance Evaluation of Swin Vision Transformer Model using Gradient Accumulation Optimization Technique,” Jul. 2023, [Online]. Available: http://arxiv.org/abs/2308.00197

[8] P. L. Bartlett, P. M. Long, G. Lugosi, and A. Tsigler, “Benign overfitting in linear regression,” *Proceedings of the National Academy of Sciences*, vol. 117, no. 48, pp. 30063–30070, Dec. 2020, doi: 10.1073/pnas.1907378117.

[9] X. Ying, “An Overview of Overfitting and its Solutions,” *J Phys Conf Ser*, vol. 1168, p. 022022, Feb. 2019, doi: 10.1088/1742-6596/1168/2/022022.

[10] T. Dietterich, “Overfitting and undercomputing in machine learning,” *ACM Comput Surv*, vol. 27, no. 3, pp. 326–327, Sep. 1995, doi: 10.1145/212094.212114.

[11] J. A. Cook and J. Ranstam, “Overfitting,” *British Journal of Surgery*, vol. 103, no. 13, pp. 1814–1814, Nov. 2016, doi: 10.1002/bjs.10244.

[12] D. M. Hawkins, “The Problem of Overfitting,” *J Chem Inf Comput Sci*, vol. 44, no. 1, pp. 1–12, Jan. 2004, doi: 10.1021/ci0342472.

[13] D. A. van Dyk and X.-L. Meng, “The Art of Data Augmentation,” *Journal of Computational and Graphical Statistics*, vol. 10, no. 1, pp. 1–50, Mar. 2001, doi: 10.1198/10618600152418584.

[14] C. Shorten and T. M. Khoshgoftaar, “A survey on Image Data Augmentation for Deep Learning,” *J Big Data*, vol. 6, no. 1, p. 60, Dec. 2019, doi: 10.1186/s40537-019-0197-0.

[15] Z. Zhong, L. Zheng, G. Kang, S. Li, and Y. Yang, “Random Erasing Data Augmentation,” *Proceedings of the AAAI Conference on Artificial Intelligence*, vol. 34, no. 07, pp. 13001–13008, Apr. 2020, doi: 10.1609/aaai.v34i07.7000.